\documentclass[runningheads]{llncs}
\usepackage{graphicx}
\input{epsf}
\begin{document}
\title{A Psychopathological Approach to Safety Engineering in AI and AGI 
 }

\titlerunning{Psychopathological Approach to AI Safety}
%
\author{Vahid Behzadan\inst{1} \and 
Arslan Munir\inst{1}\and 
Roman V. Yampolskiy\inst{2}} 

\authorrunning{V. Behzadan et al.}
%
\institute{Kansas State University, Manhattan, KS 66506, USA\\
\email{\{behzadan,amunir\}@ksu.edu}\\
\url{http://blogs.k-state.edu/aisecurityresearch/} \and
University of Louisville, Louisville, KY 40292, USA\\
\email{roman.yampolskiy@louisville.edu}\vspace{-2 mm}}
\maketitle              
\begin{abstract}
The complexity of dynamics in AI techniques is already approaching that of complex adaptive systems, thus curtailing the feasibility of formal controllability and reachability analysis in the context of AI safety. It follows that the envisioned instances of Artificial General Intelligence (AGI) will also suffer from challenges of complexity. To tackle such issues, we propose the modeling of deleterious behaviors in AI and AGI as psychological disorders, thereby enabling the employment of psychopathological approaches to analysis and control of misbehaviors. Accordingly, we present a discussion on the feasibility of the psychopathological approaches to AI safety, and propose general directions for research on modeling, diagnosis, and treatment of psychological disorders in AGI.

\end{abstract}
\vspace{-2 mm}
\section{Introduction}
\vspace{-2 mm}
While the adaptive mechanisms of human cognition provide the means for unique skills in adjusting to dynamic environments, they are also prone to psychological disorders, broadly defined as self-reconfigurations in cognition and behavior that are deleterious to the core and long-term objectives of self or the social ecosystem \cite{american2013diagnostic}. Extrapolating from this phenomenon, it is not hard to conclude that instances of Artificial General Intelligence (AGI), which aim for similar cognitive functions, may also be prone to such disorders. For instance, certain objective functions and environmental conditions may lead a Reinforcement Learning (RL) agent to develop addictive behavior through repetitive gains of high rewards from policies that contradict the long-term objectives of the agent \cite{yampolskiy2014utility}. Other instances of such emergent disorders include post-traumatic behavior, depression, and psychosis \cite{ashrafian2017can}. It is further hypothesized that behavioral disorders may emerge as higher-order consequences of unsafe inverse RL and apprenticeship learning, by adopting manifested disorders or triggering harmful cognitive traits \cite{yampolskiy2016taxonomy}.

Current research in AI safety is generally focused on safety-aware design and mitigation techniques \cite{AISafetyLandscape}, but the expanding complexity of AI and in particular AGI will render such analysis as difficult as those of biological intelligence and the corresponding disorders. To tackle such difficulties in human intelligence, the causes and dynamics of misbehaviors are studied at various levels of abstraction, ranging from neuroscience and cognitive science to psychology, psychiatry, sociology, and criminology. Inspired by the advantages of such diverse vantage points, we propose that studying the complex dynamics and mechanisms of failure in AI safety can greatly benefit from abstractions that parallel those of biological intelligence. Considering the practical aims of diagnosing and correcting misbehaviors in AGI, we believe that adopting the abstraction of psychopathology provides tractable settings that also benefit from cross-domain bodies of knowledge. Furthermore, while this approach may seem to be of lower relevance at present, we argue that the advent of deep RL, along with advances in hierarchical and transfer learning may have already laid the grounds for emergence of such disorders in AI.

The goal of this paper is to provide a technical discussion and the motivation for research on the psychopathology of AI and AGI. The remainder of this paper is organized as follows: Section \ref{Sec:Psycho} presents a broad overview of psychopathology. Section \ref{Sec:PsyAI} provides a discussion on the relevance of psychology to AI, followed by establishment of parallelisms between AI safety and psychopathology. In Section \ref{Sec:Directions}, high-level areas of research are identified and detailed. Finally, Section \ref{Sec:Conclusion} concludes the paper with remarks on broader impacts of this research.

\vspace{-5 mm}
\section{What is Psychopathology?} \label{Sec:Psycho}
\vspace{-4 mm}
Psychopathology refers to the scientific study of mental disorders, their causes, and corresponding treatments \cite{butcher2018apa}. Within this context, we adhere to American Psychiatric Association (APA)'s definition of mental disorder \cite{american2013diagnostic} as ``a psychological syndrome or pattern which is associated with distress, disability, increased risk of death, or significant loss of autonomy'' (i.e., pursuit of objectives). In psychopathology, disorders are commonly identified based on four metrics of abnormality, known as the four Ds \cite{Davis2018ConceptualizingPD}: Deviance of behaviors and emotions from the norm, Distress of the individual caused by suffering from a disorder, Dysfunctions that impair the individual’s ability to perform designated or normal functions, and the Danger of individual to self or the society.

Causes of mental disorders in humans include mixtures of those inherited through \emph{genetics} (e.g., neuroticism), \emph{developmental influences} caused by parental mistreatment, social influences (e.g., as abuse, bullying), and traumatic events, and \emph{biological influences} such as traumatic brain injury and infections \cite{american2013diagnostic}.

Various models have been developed to capture the dynamics of mental disorders and their emergence. For instance, biological psychiatry, or the \emph{medical model} \cite{kendler2012dappled}, is one that explains the causes of disorders based on changes in neurological circuitry. The \emph{social model}, on the other hand, analyzes the causes of mental disorders based on social and environmental interactions \cite{kendler2012dappled}. Currently, it is widely believed that understanding psychological disorders requires the comprehensive consideration of both biological and social factors, and hence the \emph{biopsychosocial models} are generally adopted to study such phenomena. These models broadly categorize mental disorders as either cognitive or behavioral. Cognitive disorders are those caused by abnormal functioning of the underlying cognitive mechanisms, and behavioral disorders are those that are learned through developmental, environmental, and social interactions \cite{kendler2012dappled}.

Diagnosis of mental disorders is generally based on an assessment of symptoms, signs, and impairments that constitute various types of disorders. A comprehensive framework for such assessments is that of the Diagnostic and Statistical Manual of Mental Disorders (DSM) \cite{american2013diagnostic}, published by the American Psychiatric Association (APA). This manual provides a common language and standard criteria for the classification of mental disorders. Furthermore, recent advances in machine learning have given rise to various software and algorithmic tools to facilitate enhanced accuracy in classification and diagnosis of mental disorders \cite{kelly2012intelligent}.

\vspace{-1 mm}
Treatment of mental disorders is commonly via one or a hybrid of two approaches. One is \emph{Psychotherapy}, which is a form of interpersonal intervention via a range of psychological techniques. For instance, Cognitive Behavioral Therapy (CBT) is employed to modify the patterns of thought and behavior associated with a particular disorder. \emph{Medication therapy} is the other approach, which targets the physiological components of disorders. For instance, antipsychotics commonly work by blocking D2 Dopamine receptors, thus controlling the chemical reward mechanism of the brain \cite{nordstrom1993central}.

\vspace{-5.5 mm}
\section{Psychopathology and AI Safety} \label{Sec:PsyAI}
\vspace{-3.5 mm}
Since its inception, AI has been closely connected to psychology and cognitive sciences \cite{dennett1978artificial}. This connection flows in both directions: AI researchers study biological cognition and behavior as inspiration for engineered intelligence, and cognitive scientists explore AI as a framework for synthesis and experimental analysis of theoretical ideas \cite{collins2013readings}. An instance of this interconnection is Reinforcement Learning (RL), where the computational algorithms of RL, such as Temporal Difference (TD) learning were originally inspired from the dopamine system in biological brains \cite{sutton1998reinforcement}. On the other hand, the work on TD learning has provided mathematical means of modeling the neuroscientific dynamics of dopamine cells in the brain, and has been employed to study disorders such as schizophrenia and the consequences of pharmacological manipulations of dopamine on learning \cite{montague2004computational}.
\begin{figure*}[!t]
\centering
\includegraphics[width = 4.5in, bb = 14 14 1010 764]{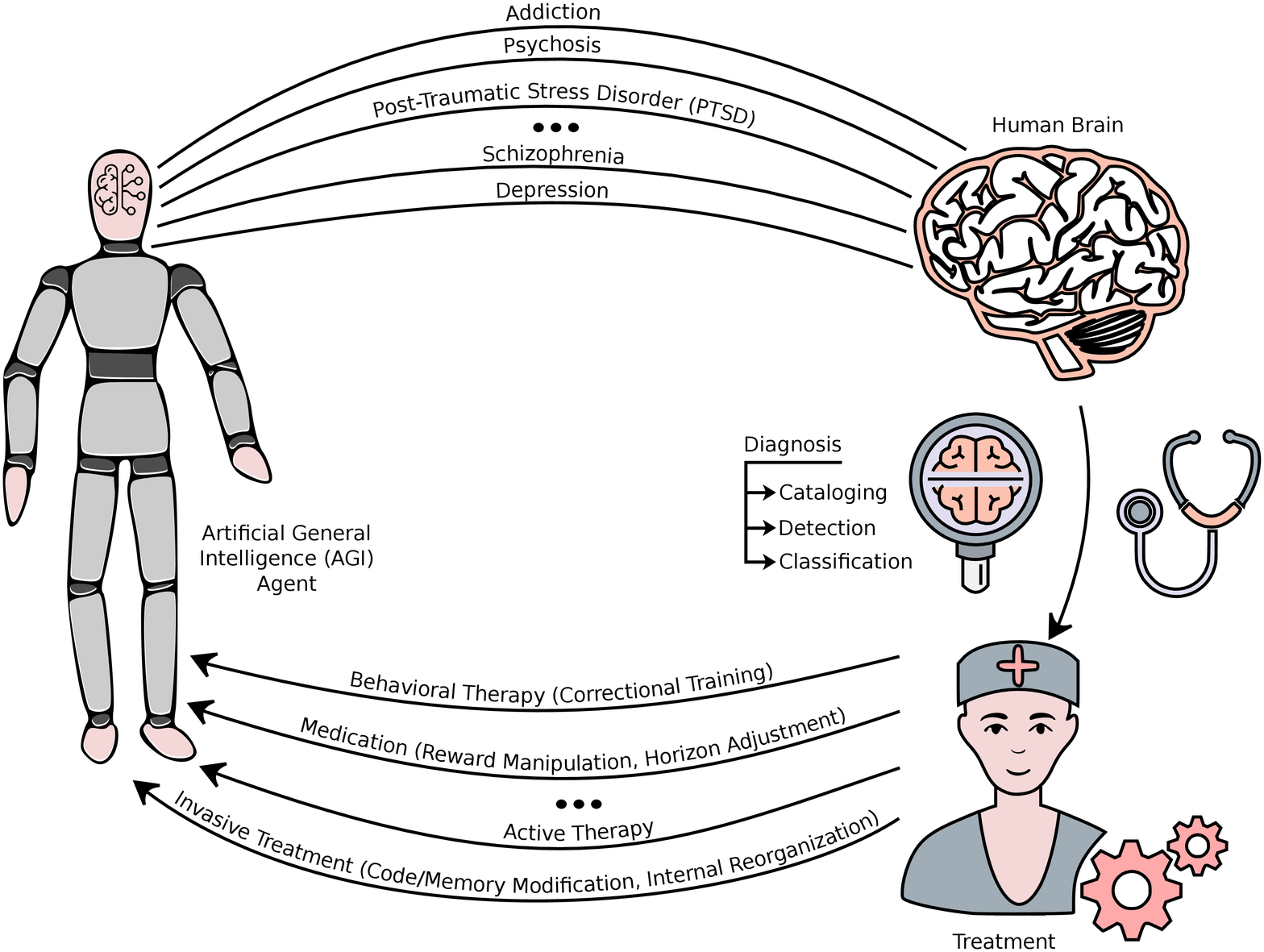}
 \vspace{-3 mm}\caption{A psychopathological approach to safety engineering in AI and AGI.}
\label{Figure_Psychopathology_AGI}
 \vspace{-7 mm}
\end{figure*}

\vspace{-1 mm}
With regards to the relationship between psychological disorders and AI safety, there are scarce and sparse resources available in the literature. Recent papers by Ashrafian \cite{ashrafian2017can} and Yampolskiy \cite{yampolskiy2014utility} \cite{yampolskiy2017detecting} present high-level arguments for the existence and emergence of mental disorders in AI. One such argument presented in \cite{ashrafian2017can} is based on the analogy of David Chalmers' philosophical zombie (p-zombie). In this analogy, the p-zombie is considered to be a fully functioning robot that acts exactly like a human-being, which is not necessarily equipped with vague notions of consciousness \cite{yampolskiy2017detecting}. The fact that this robot is capable of acting indistinguishably from humans is then used to justify that it is also prone to developmental and cognitive abnormalities that lead to misbehavior and anomalous cognition.

Furthermore, many aspects of failures in AI safety can be viewed as psychological disorders. For instance, wireheading in AI can manifest as delusional and addictive behavior. \cite{yampolskiy2014utility}. Similarly, sequences of interactions with extremely negative rewards and stresses within the exploration/exploitation trajectories of RL-based AI can potentially give rise to behavioral disorders such as depression and Post-Traumatic Stress Disorder (PTSD) \cite{ashrafian2017can}. Furthermore, the generic manifestation of the value alignment problem \cite{AISafetyLandscape} in AI is in the form of behavioral characteristics that are harmful to either the agent or the environment and society, which falls well within the definition of psychological disorders.

While \cite{ashrafian2017can} and a few other papers (e.g., \cite{atkinson2015emerging}\cite{AISafetyLandscape}) present high-level arguments on the advantages of investigating the psychopathology of AI, there remains a wide gap in satisfying the need for technical studies and practices. This paper presents a research agenda that will fill this gap via the following proposals, also illustrated in Figure~\ref{Figure_Psychopathology_AGI}.

\vspace{-5 mm}
\section{Directions of Research} \label{Sec:Directions}
\vspace{-5 mm}
Developing solid grounds for research on the Psychopathology of AI requires investigations in three main areas: Modeling and Verification, Diagnosis, and Treatment. In this section, we define and discuss the scope of each area.

\vspace{-5 mm}
\subsection{Modeling and Verification Tools}
\vspace{-3 mm}
While the descriptive similarities of human psychopathology and AI failures provide some insights into adopting such abstractions for AI safety, taking an engineering approach requires formal and mathematical modeling of the aspects and dimensions of these similarities. Such formalisms may benefit from those that already exist in the realm of cognitive and medical sciences, such as cognitive architectures \cite{kotseruba2016review} and RL-based models of the dynamics in mental disorders (e.g., \cite{montague2004computational}). Also, the quantitative analysis of such disorders necessitate the exploration and development of new models of AI and AGI based on such paradigms as neuroeconomics, complex adaptive systems, control theory, and dynamic data-driven application systems.

Furthermore, verification and validation of such models and the ensuing theories requires the development of experimental frameworks and simulation platforms. Such platforms must provide the means for wide ranges of experiments on emergence and dynamics of behavioral and cognitive disorders in arbitrary and context-dependent scenarios, and shall be compatible for various agent and environmental models.


\vspace{-5 mm}
\subsection{Diagnosis and Classification of Disorders}
\vspace{-3 mm}
This venue is on investigating and development of techniques for diagnosis of disorders in AI. Within the context of AI safety engineering, diagnosis refers to two inter-related tasks: first is to detect anomalous behaviors, and the second is to classify the type of anomalous behavior as a first step towards treatment. Detection of undesired behavior is an active topic of research in AI safety, with initial solutions such as tripwires and honeypots \cite{AISafetyLandscape} already proposed and investigated. We propose to extend current state of the art in detection through adoption and automation of parallel techniques in psychopathology. Similar to diagnostics criteria in human psychology \cite{american2013diagnostic}, a promising approach is to identify statistical deviations in behavior, as well as general indicators of misbehavior. To this end, development of machine learning approaches similar to those applied in cybersecurity for threat and intrusion detection can be a promising direction. Furthermore, generic indicators of misbehavior can be learned from models trained on simulated and annotated scenarios of disorders.

Once a misbehavior is detected, the next step is to characterize and classify the disorder that has led to such behavior. A prerequisite to this process is having a catalog of different disorders and the corresponding criteria for diagnosing such disorders. Therefore, a necessary step is the compilation of representative and experimentally verified disorders, such as addiction and anxiety in RL agents, along with manually and automatically generated criteria and characteristics of each disorder based on behavioral observations. This task shall aim to produce human- and machine-readable catalogs as AI analogues of APA’s DSM 5 \cite{american2013diagnostic}.

Besides general behavioral characteristics, there are other sources of data that can be of diagnostic value. Instances include indicators of disorders that are obtained through direct and targeted interactions with AI (similar to psychiatric evaluation of human patients), non-invasive analysis of internal states and parameters (similar to F-MRI and EEG tests of human patients), and induction or invocation of internal debug modes (similar to states of hypnosis). Exploring such ideas and approaches may greatly enhance the accuracy of diagnosis, and lead to novel techniques for psychoanalysis and diagnostics of AI and AGI.

\vspace{-5 mm}
\subsection{Treatment}
\vspace{-4 mm}
When a disorder is diagnosed in an AI agent, it is not always feasible to simply decommission or reset the agent. In such cases, it is often preferable to pursue treatment via minimally destructive techniques that correct the misbehaviors of agent, while preserving the useful traits learned by that agent. Such treatments need to satisfy a number of challenging requirements. Advanced AI are complex adaptive systems, and therefore minor perturbations of one component may lead to unintended consequences on local and global scales. For instance, correcting a developmental disorder by removing a series of harmful experiences from the memory of an AI may lead to behavioral changes that are even more undesirable than the original misbehavior. Therefore, effective treatments must either be minimally invasive or non-invasive at all.

Inspired by psychopathological parallels, we propose two general approaches to treatment of pathologies in AI. One is correctional training, which adopts the approach of behavioral therapy. This approach is to retrain an agent in controlled environments and scenarios, such that harmful experiences can be remedied or alleviated through new experiences. The second approach parallels that of medication therapy, in which the reward signals of AI agents are artificially manipulated via external means to adjust their behavioral policies. This is similar to the use of anti-depressants and anti-psychotics in treating disorders related to production and inhibition of dopamine and serotonin in human brains.

\vspace{-5 mm}
\section{Conclusion} \label{Sec:Conclusion}
\vspace{-5 mm}
This paper presents the argument that while current research in AI safety is generally focused on design and mitigation problems, the complexity of AGI will render such analysis as difficult as those that capture biological intelligence and disorders. Hence, studying the complex dynamics and mechanisms of emergent failures in AI and AGI can greatly benefit from abstractions that parallel those of biological intelligence. Considering the practical objectives of diagnosing and treating misbehaviors in AGI, we propose that psychopathological approaches provide tractable settings while benefiting from various bodies of knowledge.  Accordingly, we present a high-level research agenda that includes explorations of parallels between human and AI psychopathology, development of methodologies for diagnosis of behavioral pathologies in AI, and propose techniques for treatment of such disorders.

As the paper detailes, psychology and AI enjoy a bi-directional flow of inspirations. A major impact of the proposed research is the production of outcomes that can be of use and inspiration to current research in psychopathology and cognitive sciences. Furthermore, the results of this work may provide a deeper understanding of the safety requirements and guidelines for designing advanced AI and AGI, while guiding policy makers on the risks and potential solutions involved in the integration of AGI into societies. We hope that this paper motivates initial efforts in laying solid foundations for future research and developments in this scarcely explored but promising venue.

\vspace{-3 mm}
{\tiny

}
%
%
%
%
\end{document}